# An Axis-Based Representation for Recognition

## Cagri Aslan and Sibel Tari

*cagriaslan@yahoo.com, stari@metu.edu.tr*
*Department of Computer Engineering*
*Middle East Technical University, TR-06531*

## Abstract

*This paper presents a new axis-based shape representation scheme along with a matching framework to address the problem of generic shape recognition. The main idea is to define the relative spatial arrangement of local symmetry axes and their metric properties in a shape centered coordinate frame. The resulting descriptions are invariant to scale, rotation, small changes in viewpoint and articulations. Symmetry points are extracted from a surface whose level curves roughly mimic the motion by curvature. By increasing the amount of smoothing on the evolving curve, only those symmetry axes that correspond to the most prominent parts of a shape are extracted. The representation does not suffer from the common instability problems of the traditional connected skeletons. It captures the perceptual qualities of shapes well. Therefore finding the similarities and the differences among shapes becomes easier. The matching process gives highly successful results on a diverse database of 2D shapes.*

## 1. Introduction

Shape information plays a key role in the overall perception process. Although considerable progress has been made in its representation and matching, generic shape recognition problem remains largely unsolved. Most of the implemented representation schemes are suited for narrow domains where there is limited and predictable variability of input data. They differ from each other by the aspects of the shape that they make explicit.

Generic shape recognition demands representations that can capture the large degree of variability as a result of changes in illumination, viewpoint, rotation, scale, articulation etc. Many researchers have tried to identify the requirements of shape representation schemes that can be used for generic shape recognition e.g. [5]. The idea of decomposing a shape into primitives and building up its description in a frame that expresses the links between these primitives was first made explicit by Marr and Nishihara [6] and has been one of the most promising guidelines for recognition. Representations based on symmetry axes have been considered in this respect because of their ability to capture the perceptual properties of shapes.

An early axis-based representation in the literature is the prairie fire model of Blum [2]. The shape boundary evolves in the inward direction with a constant speed producing shocks (quench points). The locus of quench points and their time of formation define a *morphological skeleton*. Morphological skeleton is an instable representation: a small change in the shape may cause a significant change in its description.

A variety of techniques have been suggested to overcome this instability problem. Traditionally, pruning of the axes has been mostly used to regularize the morphological skeleton. Pruning methods define a saliency measure for axis points and discard those points whose significance are below a threshold. Axis length, propagation velocity, maximal thickness, the ratio of the axis and the boundary it unfolds are the most typical significance measures which do not reflect the perceptual prominence of parts well [9].

With the developments in curve evolution and the introduction of *reaction-diffusion scale space* by Kimia et al. [4], it became possible to combine skeletonization and smoothing into a single process. The amount of diffusion (smoothing) determines the detail of the skeletal description or the scale of the representation. Survival of a branch over scales is a measure of significance [11]. Though this idea of a combined framework is appealing, it has not been used in practice for obtaining stable axial descriptions for recognition. The researchers who has proposed recognition frameworks based on this formulation used only those axial descriptions obtained by morphological evolution [8,10]. This may be due to two facts. First, when diffusion is introduced, detection of *first order shocks,* which are the local curvature maxima of the evolving curve, becomes difficult. Second, even a small amount of diffusion leads to a disconnected skeleton. This is not an artifact of computation. Symmetry points measure the deviation of the evolving boundary from a circle. Hence, when a curve locally gets rid of a protrusion or an indentation -under the influence of diffusion- the symmetry branch tracking it terminates. Deriving a hierarchical representation from a disconnected skeleton is a more difficult problem. An alternative implementation is provided by Tari, Shah and Pien [12]. They introduced a surface whose level curves correspond to the smoothed fire front. Key to their work is the inverse proportionality of the level curve curvature to the surface gradient which allowed them to capture the local symmetry points even under significant amount of diffusion.

When skeletons are used for shape matching and recognition, the common paradigm is to convert the skeletal description to a graph or a tree and reduce the problem to matching of these structures. Existing methods mainly differ from each other by the distance measures they use to compute similarities between representation primitives and by the graph (or tree) matching algorithms they employ, e.g. [3,7,8,10,13]. An interesting idea in Zhu and Yuille [13] is the generation of more than one possible skeleton graph for the input shape to overcome the unreliability of the skeleton. Even though the approaches based on connected skeletons are successful to some extent, the instabilities of the representations lower their performance. Also, the complexity of the descriptions or the data structures leads to computationally expensive matching and recognition algorithms. These rich descriptions may be suitable for reconstructing a shape, but may not be necessary for recognition.

Our approach is to derive from shapes their coarsest level descriptions in the form of a disconnected set of axial branches. Relative placement of the branches and their metric properties are measured in a polar coordinate frame centered on the shape. In this respect, it is quite similar to the 3D model representation of Marr and Nishihara [6] in which the spatial arrangement of major component axes are specified by a model axis that provides coarse information. An important property of our representation is that it can produce descriptions that are variant to changes in scale, rotation, and viewpoint in addition to the descriptions that are invariant to these changes. Even though invariance to these transformations is desirable, there are situations in which transformation variant descriptors must be used, e.g. discriminating '6' from '9'.

Underlying method of symmetry point detection is closely related to the method of Tari, Shah and Pien [12]. The shape matching process is a branch and bound algorithm which searches over all possible matchings between two shapes. Even though the worst case complexity of the branch and bound algorithm is high, in practice our matching process is very fast because the number of primitives in the descriptions is small and the number of permutations that need to be tested are decreased using additional constraints.

## 2. Detection of Symmetry Axes

### 2.1 Detection of symmetry points

The symmetry point detection method is the method of Tari, Shah and Pien [12] (TSP) with the exception that we take the smoothing parameter to infinity when computing the distance surface. In TSP, the basic tool is the function $v$ whose level curves are interpreted as a family of evolving curves under the influence of constant and curvature motions. When compared to standard implementation methods [4,11], this one is much simpler and much faster. Function $v$ is computed by solving the following equation:

$$\nabla^2 v - \frac{v}{\rho^2} = 0, v \mid_\Gamma = 1$$

where $\Gamma$ is the shape boundary.

The symmetry points which track the protrusions and indentations of the evolving curve are given by the minima and maxima of the gradient along the level curve respectively. The vanishing of the gradient provides further information. They are the shape centers where the level curves of the shape shrink into a point or the break points due to presence of narrow necks. During the course of evolution a branch tracking the protrusions (a positive axis) may merge with a branch tracking the indentations (a negative axis) terminating both branches. If a branch does not terminate at such a junction, it comes to rest at a surface extrema. Figure 1 shows the function $v$ and the symmetry axes obtained from this function in the TSP framework.

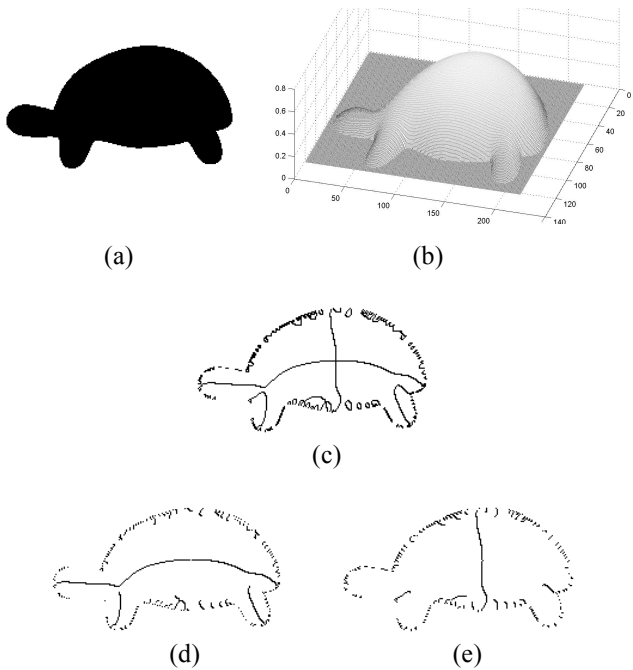

(a)                              (b)

(c)

(d)                              (e)

Figure 1. TSP Method (a) Original shape (b) Surface 1-v *(ρ = 32)* (c) Full Symmetry Points (d) Positive Symmetry Points (e) Negative Symmetry Points

In curve evolution based shape analysis, boundary smoothing is the primary method of axis regularization which alleviates most of the sensitivity problems. A scale space representation is obtained by changing the amount of smoothing. The ability to generate scale-space descriptions of shapes has been considered essential for recognition. Despite its great appeal, the idea has not been applicable in practice for a number of reasons. First of all, the scales generated are not "absolute". The selection of the same smoothing parameter for different shapes does not guarantee that these shapes will be represented at the same level of detail. This is because the survival time of symmetry axes is a local property which depends on the curvature of nearby protrusions and indentations. Moreover, skeletonization methods require skeletons to be connected so that the relations among branches can be expressed easily. If a symmetry branch doesn't connect to the main skeleton, it is discarded. The transition from one scale to the other may be accompanied by substantial changes in the skeleton structure. Because of this large change, the task to determine the correspondences between symmetry branches at different levels of detail becomes a difficult problem. Unless a method is devised to compare two shapes at the same level of detail, these scale-space representations can not be used in practice.

In our work, the derived symmetry axes need not to be connected and the final data structure for matching and recognition is neither a graph nor a tree. In order to obtain a stable shape description, we propose using a sufficiently large smoothing parameter. Consider the two vases shown in Figure 2. Interpretations at ρ = 32 (which is considerably a large diffusion parameter) are significantly different. Notice that the first surface has one saddle point and two elliptic points corresponding to the neck of the vase and the centers of the top and bottom parts respectively. The second surface, on the other hand, has one elliptic point. Slight change in the thickness of the neck led to a significant change in the interpretation of the topology.

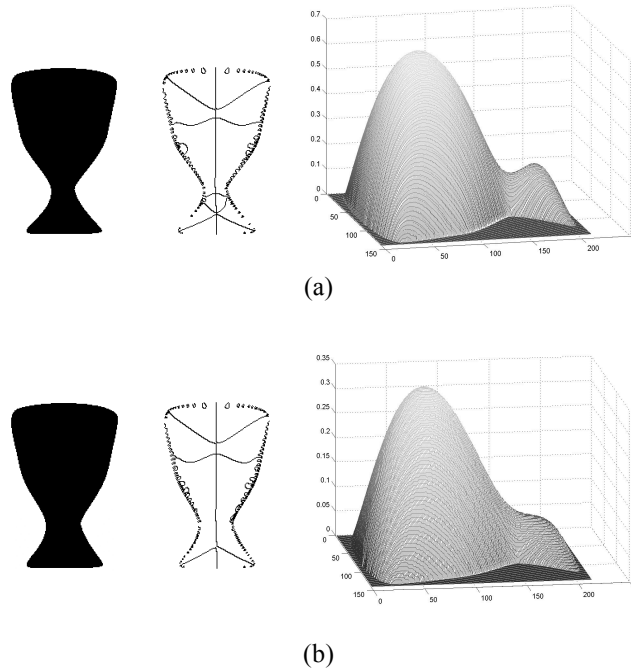

(a)

(b)

Figure 2. (a) Vase shape with a neck, its full symmetry points, and function 1-v (b) Second vase shape with a thicker neck

If our shape representation scheme is to be used on a broad shape domain where a great variability on the thickness, length, width and size is expected, the level of smoothing required for each shape should be determined. This level, which is necessary to obtain a stable description, varies from shape to shape. The computation time increases as the amount of diffusion is increased. Therefore, it is not feasible to select a very large smoothing value and use this fixed value to extract the description from all shapes. The strategy we employ is to select a small smoothing value and increase it until a function with a single extremum point is obtained which means that the description has a single center. For efficiency reasons, we use linear diffusion with Dirichlet conditions on the shape boundary.

$$\frac{\delta v}{\delta \tau} = \nabla^2 v, \; v|_\Gamma = 1$$

To describe it briefly, letting $\rho \to \infty$ drops the second term of TSP whose solution is a function which is equal to one everywhere. Yet, we can consider an iterative scheme to obtain the surface $v$ at a critical time T during its evolution towards ones everywhere. Sufficiently evolved surface has a single elliptic point corresponding to a single shape center. Details of the computation is given in our technical report [1].

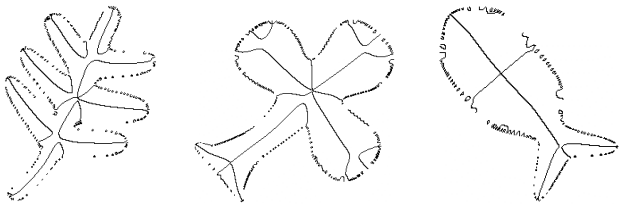

Figure 3. Full symmetry points of some shapes with significant necks.

Figure 3 shows symmetry points we compute for some shapes. Despite their obvious necks, the representation interprets them as a single blob. There is one practical difficulty associated with some dog-bone or dumbbell-like shapes where the two main parts of the shape have nearly the same prominence (Figure 4). It takes a significant amount of computation time to reduce these kinds of shapes to a single point. Therefore, it is logical to retain their dumbbell-like topology in the final description. Having two types of descriptions may lead to instability when some shapes that are between these two types are encountered. This is a trade-off between computational efficiency versus accuracy.

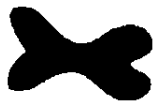

Figure 4. A dog-bone shape

## 2.2 Grouping symmetry points into axes and pruning

The next step in deriving the skeletal representation is to group symmetry points into symmetry axes. This is not a trivial task. In [11] a set of rules to group skeletal points are presented. In our work, the grouping is roughly based on the connectedness [1]. We use pruning only to get rid of the small noise branches near the boundary (discretization artifacts). Figure 5 shows some results.

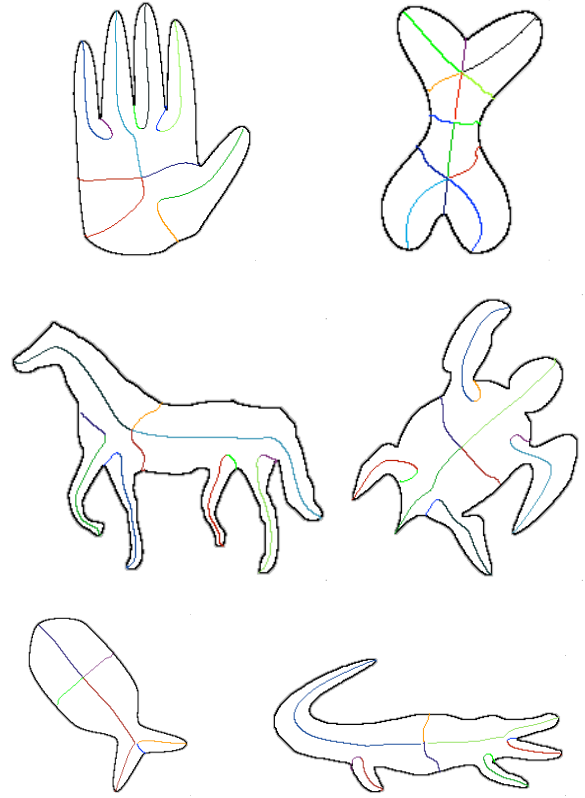

Figure 5. Full symmetry axes of some shapes after grouping and pruning.

## 3. The Representation

### 3.1 Setting up the canonical coordinate frame

If a symmetry branch survives long enough, it comes to rest at a shape center or a neck point. There are always at least two positive and two negative symmetry branches that flow into a shape center (elliptic point) [12]. These branches represent the most prominent features of the shape. During the evolution, when all minor branches have terminated at junction points, the resulting shape includes only the most significant branches and it can be considered as the coarsest description of the original shape. A shape may undergo changes in scale, rotation, and viewpoint. It may also undergo non-rigid transformations such as articulation and boundary perturbations. However, the coarsest structure will remain almost the same. Consider Figure 6. The two positive symmetry axes (red) and the two negative symmetry axes (yellow) reach the shape center (green). When all the branches except these major ones have terminated, the shape becomes, in its coarsest form, an ellipse.

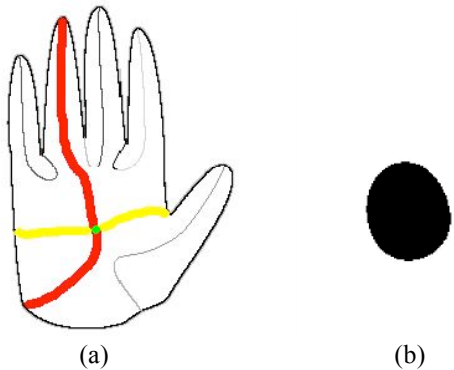

(a)                    (b)

Figure 6. (a) Most prominent branches of the hand shape. (b) The state of the hand shape at the time the branches except the major ones have terminated.

The center point and one of the major axes allows us to set up a canonical coordinate frame (Figure 7). Any one of the major axes can be selected. The line connecting the origin to a nearby point on the selected major axis defines the reference axis. This point on the major axes should be chosen within the ellipse representing the coarsest form because the major axes may bend or even bifurcate as we move away from this region (Figure 8).

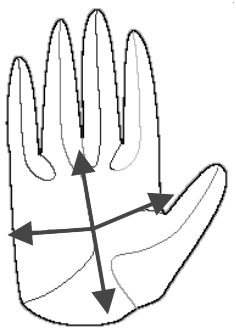

Figure 7. Four possible reference axes of the hand shape

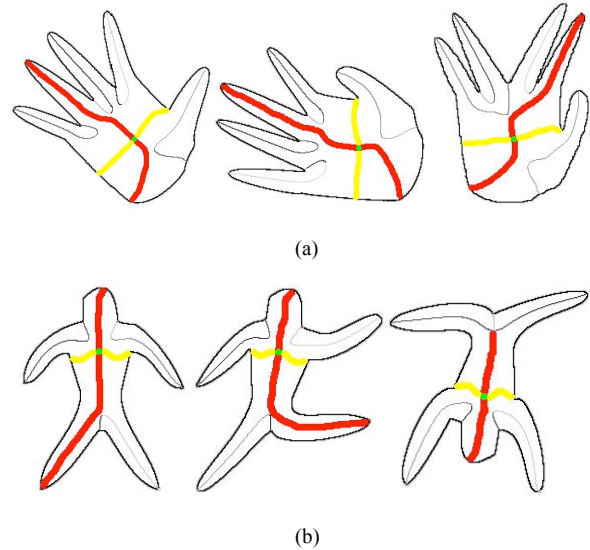

(a)

(b)

Figure 8. The major axes of (a) the hand shape and (b) a human shape.

No matter which major axis is chosen as a reference axis, the same axis must be chosen for similar shapes. Since there are two major axes of the same type, there is an ambiguity in the process. If the descriptions of two similar shapes depend on different coordinate frames, the matching algorithm will be unable to determine the similarities of shapes. This situation may necessitate creating at least two descriptions. In this paper, we employ a strategy that reduces the matching time of our branch and bound algorithm drastically. We use the two major symmetry axes of the same type and describe a shape as two halves. Each half is represented in its own coordinate frame.

For a dumbbell-like shape, one of the three surface extrema may be chosen as the origin. The fact that each hyperbolic point of the surface has at least two positive symmetry axes with negative curvature [12] removes this ambiguity (Figure 9).

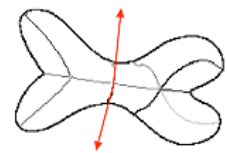

Figure 9. Reference axes for a dog-bone shape.

## 3.2    Spatial organization of symmetry branches

The relative placement of symmetry axes and their metric properties e.g. length are measured in the chosen coordinate frame. Each symmetry branch is represented by an arrow from the origin to the termination point. The termination points are used because when a shape's limbs articulate, the points where they connect to the main part of the shape tends to remain the same.

The length of the arrow defines $r$; the angle between the arrow and the reference axis defines θ providing a polar representation. Symmetry axes are added to a subshape in a counter clockwise direction, hence, the array of symmetry branches is sorted in ascending order of their angle with reference axes (Figure 10). This enables the use of an order constraint in the matching process.

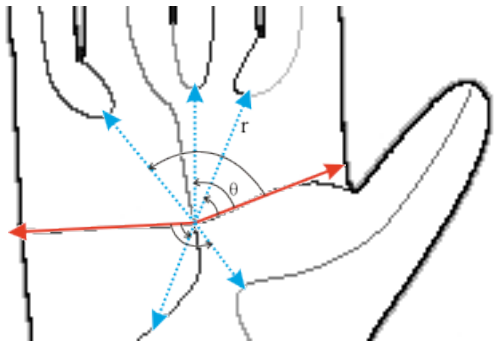

Figure 10. The reference axes (red) and the position vectors (blue) of the symmetry axes of the hand shape.

### 3.3 The canonical coordinate frame: handling ambiguities

In the previous section, we assumed that only four symmetry axes flow into the shape center. More complicated situations may occur when more than two negative branches reaches to the center. That is when the symmetry is more than two-fold. We have to guarantee that the same coordinate frame is formed for the shapes in the same class. A simple solution is to interpret these situations as the ambiguities of the representation and to generate a number of possible descriptions. If there are n major axes that reach the shape center, we select all the two permutations of n major axes to generate possible descriptions (Figure 11).

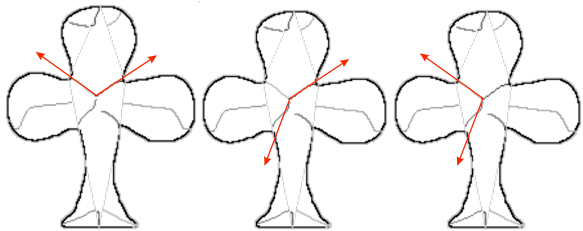

Figure 11. Possible reference axes of a shape.

This redundancy of descriptions doesn't incur high computational penalties in the matching process. First, not too many shapes have this property and typically at most three or four negative axes come near the shape center.

## 4. Shape Matching

Shape matching stage where the best correspondences between two shapes are determined, is the basis of recognition. A similarity measure is defined and the correspondences are ranked according to the similarity score they produce. In our matching framework, the local symmetry axes are the primitives of the shape description. The information stored for each primitive is summarized in Table 1.

The location in polar coordinates (r, θ), the normalized length and the type information of a branch are used to compute the similarity between two branches. It is natural to compare the features of two branches independently and obtain a similarity score based on the averaging of the similarity scores of the features. A normalized similarity scale that varies between 0 and 1 is used, with 1 indicating that the two axes are identical. If the types of branches are different, they are simply not matched. The similarity score for location features is computed using:

$$f_{sim}^i \left( f_0^i, f_1^i \right) = 1 - \max \left( \left| f_0^i - f_1^i \right| / f_{thr}^i, 0 \right), i = r, \theta$$

The score for normalized length is computed similarly except that it is also normalized by the total axes length. If any $f^i{}_{sim}$ is zero, the score is considered to be zero. This prevents the matching of two branches that are very different in some aspects but similar in others.

The order of the branches along the shape boundary is also stored in the description. It is used to sort out impossible correspondences in the matching process. This reduces computation time and leads to perceptually more accurate matchings.

The proposed description is invariant to scale, rotation and translation. Some applications may not desire invariance, thus we store the extrinsic coordinates of the center point (for translation), the total length of symmetry axes (for scale) and the orientation of the reference axes in the 2D image plane (for rotation).

| Description element | Information Stored |
|---|---|
| Shape | Center Point $(x_0, y_0)$ |
|  | Total Length of Axes |
|  | Orientation of Reference Axis $\{m_0, m_1\}$ |
| Local Symmetry Axis | Type (Positive, Negative) |
|  | Location $(r, \theta)$ |
|  | Normalized Length |
|  | Reference Axis (Yes, No) |
|  | Next Symmetry Axis |
|  | Previous Symmetry Axis |

Table 1. The information stored in the descriptions

The total similarity of two shapes is determined by the weighted sum of the similarity scores of the matched branch pairs. The lengths of branches determine weights. Therefore, a prominent branch that is not matched affects the overall similarity score of the shapes significantly.

The matching process is a branch and bound algorithm that searches over all possible matchings of two shapes. The worst case complexity of this type of algorithm is high, but in practice our matching process is very fast. The number of shape primitives is small and additional measures are employed to reduce the number of permutations that need to be tested. Those matchings that would violate the order constraint are not tested. The generation of a permutation is stopped when it is determined that the current branch of computation will not be able to produce a higher similarity value than the current maximum. Lastly, the representation of the shape as two halves in two different coordinate frames makes it possible to reduce the problem into two half problems providing a drastic decrease in computation time.

## 5. Examples

We demonstrate the correspondence matching results on a few illustrative shape pairs. In Figure 12, the matching process is able to find the correspondences when a shape undergoes rotation and articulation. In the case of missing parts (Figure 13), the perceptually correct correspondences are found since the spatial organization of the symmetry branches are stored in the descriptions. The unmatched finger lowers the total similarity score significantly. In Figure 14, the matching of a horse and a cat yields a similarity value of 0.711. The differences in the metric properties of matched branch pairs are reflected in the overall similarity score.

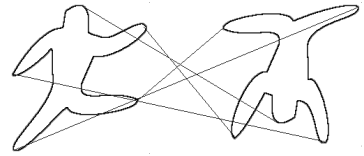

Figure 12. Similarity value is 0.918.

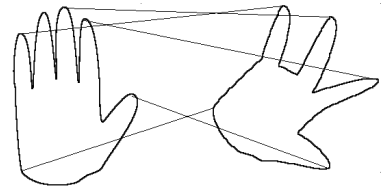

Figure 13. The similarity value is 0.728.

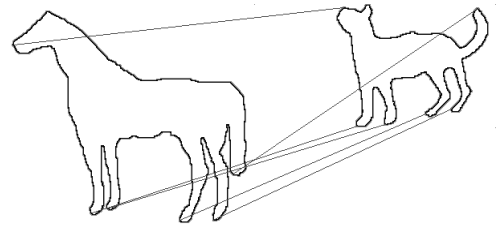

Figure 14. The similarity value is 0.711.

We have evaluated the recognition performance of our system on a diverse shape database. As shown in Figure 15, the database includes 14 categories with 4 shapes in each category. Among the shapes within the same category there are differences in orientation, scale, articulation and small boundary details. This is mainly to evaluate the performance of the matching process under visual transformations. Figure 16 shows the nearest neighbors of some query shapes. Notice that in all the example queries the top four matches are from the same category resulting in a %100 recognition rate.

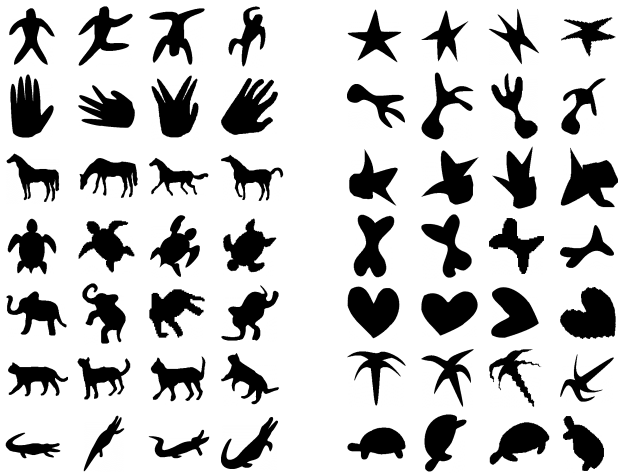

Figure 15. Our shape database

| | | | | | | |
|---|---|---|---|---|---|---|
| 0.920 | 0.915 | 0.898 | 0.892 | 0.672 | 0.653 | |
| 0.957 | 0.864 | 0.787 | 0.765 | 0.624 | 0.572 | |
| 0.923 | 0.903 | 0.796 | 0.752 | 0.724 | 0.703 | |
| 0.912 | 0.906 | 0.874 | 0.852 | 0.763 | 0.705 | |
| 0.887 | 0.887 | 0.803 | 0.827 | 0.680 | 0.671 | |
| 0.892 | 0.841 | 0.805 | 0.646 | 0.500 | 0.483 | |
| 0.837 | 0.809 | 0.803 | 0.610 | 0.580 | 0.549 | |
| 0.890 | 0.877 | 0.835 | 0.824 | 0.824 | 0.796 | |

Figure 16. Some query results

## 6. Conclusion

We presented a non-conventional approach to shape recognition using skeletons. Unlike common skeletal representations, our branches are disconnected. It is precisely the disconnected nature of the branches that enables us to define a shape centered reference frame and measure metric properties easily and accurately. In this representation, both the invariant and variant interpretations can be generated quite robustly. Use of extremely large regularization value is the major source of robustness. In fact, it is the regularization which leads to disconnectedness. Proposed matching algorithm is able to find the perceptually correct correspondences and produce a similarity score which may be interpreted as a probability of shape equivalence and may be used as an index in very large shape databases.